\documentclass[runningheads,a4paper]{llncs}
\usepackage{url}
\urldef{\mailsa}\path|{alfred.hofmann, ursula.barth, ingrid.haas, frank.holzwarth,|
\urldef{\mailsb}\path|anna.kramer, leonie.kunz, christine.reiss, nicole.sator,|
\urldef{\mailsc}\path|erika.siebert-cole, peter.strasser, lncs}@springer.com|    
\newcommand{\keywords}[1]{\par\addvspace\baselineskip
\noindent\keywordname\enspace\ignorespaces#1}
\usepackage{enumitem}
\usepackage{amsfonts}
\usepackage{amsmath}
\usepackage{float}

\usepackage{times}
\usepackage{graphicx,subfigure}
\usepackage{pgfplots}
\usepackage{pgfplotstable}
\usepgfplotslibrary{groupplots}
\usepackage{tikz}

\usepackage{bm}
\usepackage{algorithm}
\usepackage{algorithmic}

\usepackage[hang]{footmisc}
\usepackage{lipsum}
\usepackage{hyperref}
\usepackage{mathtools}

\usepackage[nocompress]{cite}

\newcommand{\vect}[1]{\mathbf{#1}}
\newcommand{\real}[1]{\mathbb{#1}}
\newcommand{\forward}[1]{\overrightarrow{#1}}
\newcommand{\backward}[1]{\overleftarrow{#1}}

\newcommand{\PSE}{\texttt{PSE}}
\newcommand{\PW}{\texttt{PW}}
\newcommand{\AEP}{\texttt{AEP}}
\newcommand{\AQ}{\texttt{AQ}}

\newcommand{\OLD}{\texttt{OLD}}

\newcommand{\NAB}{\texttt{NAB}}

\begin{document}

\mainmatter  

\title{Position-based Content Attention for Time Series Forecasting with Sequence-to-sequence RNNs}

\titlerunning{Position-based Content Attention for Time Series Forecasting}

%
%
\author{Yagmur Cinar$^*$ \and Hamid Mirisaee$^*$ \and Parantapa Goswami$^+$ \and Eric Gaussier$^*$ \and Ali A\"{i}t-Bachir$^\dagger$ \and Vadim Strijov$^\ddag$}
\authorrunning{Y. Cinar, H. Mirisaee, P. Goswami, E. Gaussier, A. A\"{i}t-Bachir, V. Strijov}

\institute{$^*$ Univ. Grenoble Alpes, CNRS, Grenoble INP, LIG\\
$^+$ Viseo R\&D, $^\dagger$ Coservit, $\ddag$ CCRAS\\
$^*$\url{firstname.lastname@imag.fr}, $\ddag$ \url{Strijov@ccas.ru}\\
$^+$\url{parantapa.goswami@viseo.com}, $^\dagger$\url{a.ait-bachir@coservit.com}\\}

%
%

\toctitle{Lecture Notes in Computer Science}
\tocauthor{Authors' Instructions}
\maketitle

\begin{abstract}
We propose here an extended attention model for sequence-to-sequence recurrent neural networks (RNNs) designed to capture (pseudo-)periods in time series. This extended attention model can be deployed on top of any RNN and is shown to yield state-of-the-art performance for time series forecasting on several univariate and multivariate time series.
\end{abstract}

\section{Introduction}
\label{sec:intro}
\vspace{-0.2cm}
Predicting future values of temporal variables is termed as {\it time series forecasting} and has applications in a variety of fields, as finance, economics, meteorology, or customer support center operations. Time series often display pseudo-periods, \textit{i.e.} time intervals at which there is a strong correlation, positive or negative, between the values of the times series. In a forecasting scenario, the pseudo-periods correspond to the difference between the positions of the output being predicted and specific inputs. Pseudo-periods may be due to seasonality or to the patterns underlying the activities measured.

A considerable number of stochastic \cite{journal/ijf/GooijeraH06} and machine learning based \cite{conf/ebiss/BontempiTB12} approaches have been proposed for this problem. A particular class of approaches that has recently received much attention for modelling sequences is based on sequence-to-sequence Recurrent Neural Networks (RNNs) \cite{arxiv/Graves13}, hereafter referred to as Seq-RNNs. In order to capture pseudo-periods in Seq-RNNs, one needs a \textit{memory} of the input sequence, \textit{i.e.} a mechanism to reuse specific (representations of) input values to predict output values. As the input sequence is usually longer than the pseudo-periods underlying the time series, longer-term memories that store information pertaining to past input sequences (as described in \textit{e.g.} \cite{WestonMemNN,WestonMemN2N,TuringGraves}) are not required. A particular model of interest here is the content attention model proposed in \cite{arxiv/BahdanauCB2014} and described in Section~\ref{sec:model-univ}. This model allows one to reuse the content of the input sequence to predict the output values. However, this model was designed for text translation and does not directly capture position-based pseudo-periods in time series. It has nevertheless been specialized in~\cite{VinyalsPointer}, under the name pointer network, so as to select the best input to be reused as the output. This model would be perfect for noise-free, truly periodic times series. In practice, however, times series are noisy and if the output is highly correlated to the input corresponding to the pseudo-period, it is not an exact copy of it.  We propose in this paper extensions of the attention model that capture pseudo-periods and lead to state-of-the-art methods for time series forecasting.

The remainder of the paper is organised as follows: Section~\ref{sec:rel-work} discusses the related work. Section~\ref{sec:model-univ} presents the position-based content attention models for both univariate and multivariate time series. Experiments illustrating the behaviour of the proposed models are described in Section~\ref{sec:exps}. Lastly, Section~\ref{sec:conclusion} concludes the paper.
\vspace{-0.6cm}

\section{Related Work}
\label{sec:rel-work}
\vspace{-0.3cm}
Various stochastic models have been developed for time series modeling and forecasting. Notable among these are autoregressive (AR) \cite{inproceedings/rsl/Walker31} and moving averages (MA) \cite{journal/econo/Slutsky37} models, that were combined in a more general and effective framework, known as autoregressive moving average (ARMA), or autoregressive integrated moving average (ARIMA) when the differencing is included in the model \cite{article/as/BoxJ68}.  Vector ARIMA, or VARIMA \cite{journal/jas/TiaoB81}, is the multivariate extension of the univariate ARIMA model. More recently, based on the development of statistical machine learning, time series prediction has been formulated as a regression problem typically solved with Support Vector Machines (SVM) \cite{journal/ieeecim/SapankevychS09} and, even more recently, with Random Forests (RF) \cite{journal/fea/CreamerF04,journal/ema/KusiakVW13}. RF have been in particular used for prediction in the field of finance \cite{journal/fea/CreamerF04} and bioinformatics \cite{journal/ema/KusiakVW13}, and have been shown to outperform ARIMA in different cases \cite{Kane-2014}.

In this study, RNNs are used for modeling time series as they incorporate contextual information from past inputs and are thus an attractive choice for predicting sequence data, including time series \cite{arxiv/Graves13}. Early work \cite{conf/nips/ConnorAM91} has shown that RNNs (a) are a type of nonlinear autoregressive moving average (NARMA) model and (b) outperform feedforward networks and various types of linear statistical models on time series. Subsequently, various RNN-based models were developed for different time series, as noisy foreign exchange rate prediction \cite{journal/ml/GilesLT01}, chaotic time series prediction in communication engineering \cite{journal/science/JaegarH04} or stock price prediction \cite{journal/HsiehaHY11}. A detailed review can be found in \cite{journal/prl/LangkvistKL14} for different time series prediction tasks.

RNNs based on LSTMs \cite{journals/neco/HochreiterS97}, that we consider here, alleviate the {\it vanishing gradient} problem of the traditional RNNs. They have furthermore been shown to outperform traditional RNNs on various temporal tasks \cite{conf/icann/GersES01,journal/jmlr/GersSS2003}. Recently, they have been used for predicting the next frame in a video and for interpolating intermediate frames \cite{arxiv/RanzatoSBMCC14}, for forecasting the future rainfall intensity in a region \cite{conf/nips/ShiCWYWW15}, or for modeling clinical data of multivariate time series \cite{arxiv/LiptonKEW15}. The attention model in Seq-RNNs \cite{arxiv/BahdanauCB2014,arxiv/Graves13} has been studied very recently for time series prediction \cite{journal/jmlr/Riemer2016} and classification \cite{conf/nips/Edward2016}. In particular, the study in \cite{journal/jmlr/Riemer2016} uses the attention to determine the importance of a factor for prediction.

None of the previous studies, to the best of our knowledge, investigated the possibility to capture pseudo-periods in time series via the attention model. This is precisely the focus of the present study, that introduces generalizations of the content based attention model to capture pseudo-periods and improve forecasting in time series.

\vspace{-0.5cm}
\section{Theoretical framework}
\label{sec:model-univ}

\vspace{-0.3cm}
We first focus on univariate time series. As mentioned before, time series forecasting consists in predicting future values from past, observed values. The time span of the past values, denoted by $T$, is termed as \textit{history}, whereas the time span of the future values to be predicted, denoted by $T'$, is termed as \textit{forecast horizon} (in multi-step ahead prediction, which we consider here, $T'>1$). The prediction problem can be formulated as a regression-like problem where the goal is to learn the relation $\vect{y} = r(\vect{x})$ where $\vect{y} = (y_{T+1}, \ldots, y_{T+i}, \ldots, y_{T+T'})$ is the output sequence and $\vect{x} = (x_1, \ldots, x_j, \ldots, x_{T})$ is the input sequence. Both input and output sequences are ordered and indexed by time instants. 
For clarity's sake, and without loss of generality, for the input sequence $\vect{x} = (x_1, \ldots, x_j, \ldots, x_T)$, the output sequence $\vect{y}$ is rewritten as $\vect{y} = (y_1, \ldots, y_i, \ldots, y_{T'})$.

\subsection{Background}

Seq-RNNs with memories rely on three parts: one dedicated to encoding the input, and referred to as \textit{encoder}, one dedicated to generating the output, and referred to as \textit{decoder}, and one dedicated to the \textit{memory model}, the role of which being to provide information from the input to generate each output element. The encoder represents each input $x_j$, $1 \leq j \leq T$ as a {\it hidden state}: $\forward{\vect{h}_j} = F(x_j, \forward{\vect{h}}_{j-1})$, with $\forward{\vect{h}_j}\in\real{R}^n$ and where the function $F$ is non-linear transformation that takes different forms depending on the RNN considered. We use here LSTMs with peephole connections as described in \cite{journal/jmlr/GersSS2003}.
%
%
The function $F$ is further refined, in bidirectional RNNs \cite{journal/ieeesp/SchusterP97}, by reading the input both forward and backward, leading to two vectors $\forward{\vect{h}}_j = f(x_j, \forward{\vect{h}}_{j-1})$ and $\backward{\vect{h}}_j = f(x_j, \backward{\vect{h}}_{j+1})$. The final hidden state for any input $x_j$ is constructed simply by concatenating the corresponding forward and backward hidden states, \textit{i.e.} $\vect{h}_j = [\forward{\vect{h}}_j;\backward{\vect{h}}_j]^{\mathsf{T}}$, where now $\vect{h}_j \in \real{R}^{2n}$. 

The decoder parallels the encoder by associating each output $y_i, 1 \le i \le T'$ to a hidden state vector $\vect{s}_i$ that is directly used to predict the output:
\[
y_i = \vect{W}_{\mbox{out}} \vect{s}_i + b_{\mbox{out}}, \, s_i = G(y_{i-1},\vect{s}_{i-1},\vect{c}_i)
\]
with $\vect{s}_i \in \mathbb{R}^n$. $c_i$ is usually referred to as a \textit{context} and corresponds to the output of the memory model. In this study, the function $G$ corresponds to an LSTM with peephole connections integrating a context \cite{arxiv/Graves13}.

The memory model builds, from the sequence of input hidden states $\vect{h}_j, 1 \le j \le T$, the context vector $\vect{c} = q(\{\vect{h}_1, \ldots, \vect{h}_j, \ldots, \vect{h}_T\})$ that provides  a summary of the input sequences to be used for predicting the output. In its most simple form, the function $q$ just selects the last hidden state \cite{arxiv/Graves13}: $q(\{\vect{h}_1, \ldots, \vect{h}_j, \ldots, \vect{h}_T\}) = \vect{h}_T$. More recently, in \cite{arxiv/BahdanauCB2014}, a content attention model is used to construct different context vectors (also called attention vectors) $\vect{c}_i$ for different outputs $y_i$ ($1 \le i \le T'$) as a weighted sum of the hidden states of the encoder representing the input history: 
\begin{equation}
e_{ij} = \vect{v}_a^\mathsf{T} \tanh(\vect{W}_a \vect{s}_{i-1} + \vect{U}_a \vect{h}_j), \alpha_{ij} = \mbox{softmax}(e_{ij}), \vect{c}_i = \sum_{j=1}^{T} \alpha_{ij} \vect{h}_j
\label{eq:attention}
\end{equation}
where $\mbox{softmax}$ normalizes the vector $e_i$ of length $T$ to be the attention mask over the input. The weights $\alpha_{ij}$, referred to as the attention weights, correspond to the importance of the input at time $j$ to predict the output at time $i$. They allow the model to concentrate, or {\it put attention}, on certain parts of the input history to predict each output. Lastly, $\vect{W}_a$, $\vect{U}_a$ and $\vect{v}_a$ are trained in conjunction with the entire encoder-decoder framework.

We present below two extensions for univariate time series to integrate pseudo-periods.

\subsection{Position-based content attention mechanism}

We assume here that the pseudo-periods of a time series lie in the set $\{1, ...,T\}$ where $T$ is the history size of the time series\footnote{This assumption is easy to satisfy by increasing the size of the history if the pseudo-periods are known or by resorting to a validation set to tune $T$.}. One can then explicitly model all possible pseudo-periods as a real vector, which we will refer to as $\boldsymbol{\pi}^{(1)}$, of dimension $T$, whose coordinate $j$ encodes the importance of the input at position $j$ in the input sequence to predict output at position $i$. From this, one can modify the weight of the original attention mechanism relating input $j$ to output $i$ as follows:
\begin{equation}
e_{ij} = \left\{
\begin{aligned}
&\vect{v}_a^\mathsf{T} \tanh(\vect{W}_a \vect{s}_{i-1} + <\boldsymbol{\pi}^{(1)},\boldsymbol{\Delta}^{(i,j)}> \vect{U}_a \vect{h}_j) & \mbox{if} \, (i+T-j) \le T \nonumber \\
&0 & \mbox{otherwise} \nonumber
\end{aligned}
\right.
\label{eq:attention-tau1-orig}
\end{equation}
where $<.,.>$ denotes the scalar product and $\boldsymbol{\Delta}^{(i,j)} \in \mathbb{R}^{T}$ is a binary vector that is $1$ on dimension $(i+T-j)$ and $0$ elsewhere. $\boldsymbol{\Delta}^{(i,j)}$ thus selects the coordinate of $\boldsymbol{\pi}^{(1)}$ corresponding to the difference in positions between input $j$ and output $i$. This coordinate is then used to increase or decrease the importance of the hidden state $h_j$ in $e_{ij}$. Note that, as the history is limited to $T$, there is no need to consider dependencies between an input $j$ and an output $i$ that are distant by more than $T$ time steps (hence the test: $i+T-j \le T$).

\vspace{-1.0cm}
\begin{figure*}[ht]
	\centering
	\includegraphics[height=0.39\textwidth]{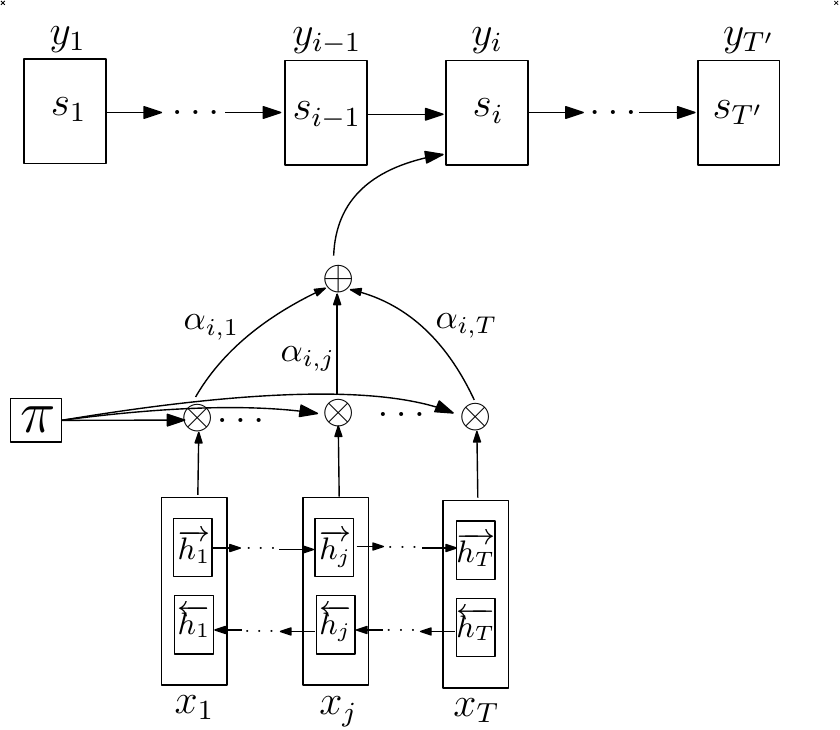} \vspace{-0.40cm}
	\caption{The illustration of the proposed position-based content attention mechanism.} \vspace{-0.5cm}
	\label{fig:attn_pi}
\end{figure*}
\vspace{-0.1cm}
The vector $e_{i}$ can then be normalized using the $\mbox{softmax}$ operator again, and a context be built by taking the expectation of the hidden states over the normalized weights. For practical purposes, however, one can simplify the above formulation by extending the vectors $\boldsymbol{\pi}^{(1)}$ and $\boldsymbol{\Delta}^{(i,j)}$ with $T'$ dimensions that are set to 0 in $\boldsymbol{\Delta}^{(i,j)}$, and by considering a vector $\boldsymbol{\Delta}$ of dimension $(T+T')$ that has $1$ on its first $T$ coordinates and $0$ on the last $T'$ ones. The resulting position-based attention mechanism then amounts to:
\begin{equation}
\mbox{RNN-}\pi^{(1)}:
\left\{
\begin{aligned}
&e_{ij} =\vect{v}_a^\mathsf{T} \tanh(\vect{W}_a \vect{s}_{i-1} + <\boldsymbol{\pi}^{(1)},\boldsymbol{\Delta}^{(i,j)}> \vect{U}_a \vect{h}_j) \boldsymbol{\Delta}_{i+T-j}\\
&\alpha_{ij} = \mbox{softmax}(e_{ij}), \, \vect{c}_i = \sum_{j=1}^{T} \alpha_{ij} \vect{h}_j
\end{aligned}
\right.
\label{eq:attention-pi1}
\end{equation}
As one can note, $\boldsymbol{\pi}^{(1)}_{i+T-j}$ will either decrease or increase the hidden state vector $\vect{h}_j$ for output $i$. Since $\boldsymbol{\pi}^{(1)}$ is learned along with the other parameters of the Seq-RNN, we expect that $\boldsymbol{\pi}^{(1)}_{i+T-j}$ will be high for those values of $i+T-j$ that correspond to pseudo-periods of the time series.
We will refer to this model as RNN-$\pi^{(1)}$.
Lastly, note that the original attention mechanism can be recovered by setting $\boldsymbol{\pi}^{(1)}$ to $\vect{1}$ (a vector consisting of $1$ on each coordinate).

In the above formulation, the position information is used to modify the importance of  each hidden state in the input side. It may be, however, that some elements in $\vect{h}_j$ are less important than others to predict output $i$. It is possible to capture this by considering that, instead of having a scalar at each position relating the input to the output, one has a vector in $\mathbf{R}^{2n}$ that can now reweigh each coordinate of $\vect{h}_j$ independently. This leads to:
\begin{equation}
\mbox{RNN-}\pi^{(2)}:
\left\{
\begin{aligned}
&e_{ij} =\vect{v}_a^\mathsf{T} \tanh(\vect{W}_a \vect{s}_{i-1} + \vect{U}_a ((\boldsymbol{\pi}^{(2)} \boldsymbol{\Delta}^{(i,j)}) \, \odot \, \vect{h}_j)) \boldsymbol{\Delta}_{i+T-j}\\
&\alpha_{ij} = \mbox{softmax}(e_{ij}), \, \vect{c}_i = \sum_{j=1}^{T} \alpha_{ij} \vect{h}_j
\end{aligned}
\right.
\label{eq:attention-pi1}
\end{equation}
where $\odot$ denotes the Hadamard product (element wise multiplication) and $\boldsymbol{\pi}^{(2)}$ is a matrix in $\mathbf{R}^{2n \times (T+T')}$. $\boldsymbol{\Delta}^{(i,j)}$ and $\boldsymbol{\Delta}$ are defined as before. We will refer to this model as RNN-$\pi^{(2)}$.

Figure~\ref{fig:attn_pi} illustrates the overall network in which $\boldsymbol{\pi}$ is a vector for $\mbox{RNN-}\pi^{(1)}$ and a matrix for  $\mbox{RNN-}\pi^{(2)}$.  

\vspace{-0.3cm}

\subsection{Multivariate Extensions}

As each variable in a $K$ multivariate time series can have its own pseudo-periods, a direct extension of the above approaches to multivariate time series is to consider that each variable $k, \, 1 \le k \le K$, of the time series has its own encoder and attention mechanism. The context vector for the $i^{\text{th}}$ output of the $k^{\text{th}}$ variable is then defined by $\vect{c}^{(k)}_i = \sum_{j=1}^{T} \alpha^{(k)}_{ij} \vect{h}^{(k)}_j$, where $\vect{h}^{(k)}_j$ is the input hidden state at time stamp $j$ for the $k^{\text{th}}$ variable and $\alpha^{(k)}_{ij}$ are the weights given by the attention mechanism of the $k^{\text{th}}$ variable. To predict the output while taking into account potential dependencies between different variables, one can simply concatenate the context vectors from the different variables into a single context vector $\vect{c}_i$ that is used as input to the decoder, the rest of the decoder architecture being unchanged:
\[
\vect{c}_i = [ \vect{c}^{(1)^T}_i \cdots \vect{c}^{(K)^T}_i]^\mathsf{T}
\]
As each $\vect{c}^{(k)}_i$ is of dimension $2n$ (that is the dimension of the input hidden states), $\vect{c}_i$ is of dimension $2Kn$. This strategy can readily be applied to the original attention mechanism as well as the ones based on $\boldsymbol{\pi}^{(1)}$ and  $\boldsymbol{\pi}^{(2)}$.

It is nevertheless possible to rely on a single attention model for all variables while having separate representations for them in order to select, for each output, specific hidden states from the different variables. To do so, one can simply concatenate the hidden states of each variable into a single hidden state ($\vect{h}_j =  [\vect{h}^{(1)^T}_j \cdots \vect{h}^{(K)^T}_j]^\mathsf{T}$) and deploy the previous attention model on top of them. This leads to the multivariate model which we refer to as RNN-$\pi^{(3)}$ and is based on the same ingredients and equations as RNN-$\pi^{(2)}$, the only difference being that RNN-$\pi^{(3)}$ is now a matrix in $\mathbf{R}^{2Kn \times (T+T')}$.

We now turn to the experimental validation of the proposed models.
\section{Experiments}\label{sec:exps}
\vspace{-0.2cm}

We retained six widely used and publicly available \cite{datasets} datasets, described in Table~\ref{tab:datasets}, to assess the models we proposed. The values for the history size were set so as they encompass the known periods of the datasets. They can also be tuned by cross-validation if one does not want to identify the potential periods by checking the autocorrelation curves. In general, the forecast horizon should reflect the nature of the data and the application one has in mind, with of course a trade off between long forecast horizon and prediction quality. For this purpose, the forecast horizons of these sets along the sampling rates are chosen as illustrated in Table~\ref{tab:datasets}. All datasets were split by retaining the first 75\% of each dataset for training-validation and the last 25\% for testing. For RNN-based methods, the training-validation sets were further divided by retaining the first 75\% for training (56.25\% of the data) and the last 25\% for validation (18.75\% of the data). For the baseline methods, we used 5-fold cross-validation on the training-validation sets to tune the hyperparameters. Lastly, linear interpolation was used whenever there are missing values in the time series\footnote{We compared several methods for missing values, namely linear, non-linear spline and kernel based Fourier transform interpolation as well as padding for the RNN-based models. The best reconstruction was obtained with linear interpolation, hence its choice here.}.
\vspace{-0.5cm}
\begin{table}[h]
\centering
\caption{Datasets.}
\label{tab:datasets}
\begin{tabular}{l | c | c | c | c | c}
\hline
Name       & Usage          & \#Instances &    History  &     Forecast horizon & Sampling rate \\ \hline \hline
Polish Electricity (\PSE)      &	Univariate     &     46379  &     96 &     4  &   2 hours  \\
Polish Weather (\PW)    & Univariate          &      4595 &     548  &     7   &   1 days  \\ 
Numenta Benchmark (\NAB)       & Univariate      &     18050   &   72    &    6  & 5 minutes  \\
Air Quality (\AQ)         & Univ./Multiv.        &      9471  &     192  &     6   &  1 hour \\
Appliances Energy Pred. (\AEP)       & Univ./Multiv.      &      19735    &   216    &     6  & 10 minutes  \\
Ozone Level Detection (\OLD)       & Univ./Multiv.      &      2536    &   548    &     7  & 1 day  \\ \hline
\end{tabular}
\end{table}
\vspace{-0.75cm}

We compared the methods introduced before, namely RNN-$\pi^{(1/2/3)}$, with the original attention model (RNN-A) and several baseline methods, namely ARIMA, an ensemble learning method (RF) and the standard support vector regression methods. Among these baselines, we retained ARIMA and  RF as these were the two best performing methods in our datasets. These methods, discussed in Section~\ref{sec:rel-work}, have also been shown to provide state-of-the-art results on various forecasting problems (\textit{e.g.} \cite{Kane-2014}). For ARIMA, we relied on the seasonal variant \cite{Hyndman2008}. To implement the RNN models, we used theano\footnote{\url{http://deeplearning.net/software/theano/}} and Lasagne \footnote{\url{https://lasagne.readthedocs.io}} on a Linux system with 256GB of memory and 32-core Intel Xeon @2.60GHz. All parameters are regularized and learned through stochastic backpropagation (the mini-batch size was set to 64) with an adaptive learning rate for each parameter \cite{KingmaB14}, the objective function being the Mean Square Error (MSE) on the output. For tuning the hyperparameters, we used a grid search over the learning rate, the regularization type and its coefficient, and the number of units in the LSTM and attention models. The values finally obtained are $10^{-3}$ for the initial learning rate and $10^{-4}$ for the coefficient of the regularization, the type of regularization selected being $L_2$. The number of units vary among the set \{128, 256\} for LSTMs and \{256, 512\} for the attention models respectively. We report hereafter the results with the minimum MSE on the test set.
For evaluation, we use MSE and the symmetric mean absolute percentage error (SMAPE). MSE corresponds to the objective function used to learn the model. SMAPE presents the advantage of being bounded and represents a scaled $L_1$ error.

\vspace{0.1cm}
\noindent \textbf{Overall results on univariate time series} 
\vspace{0.1cm}

For univariate experiments using multivariate time series, we chose the following variables from the datasets: for \PW, we selected the \emph{max temperature} series from the Warsaw metropolitan area that covers only one weather recording station; for \AQ\, we selected {\emph{C6H6(GT)}}; for \AEP\, we selected the \emph{outside humidity (RH6)}; for \NAB\, we selected the \emph{Amazon Web Services CPU usage} and for \OLD\ we selected \emph{T3}. Table \ref{table:univar} displays the results obtained with the MSE (left value) and the SMAPE (right value) as evaluation measures. Once again, one should note that MSE was the metric being optimized. For each time series, the best performance among all methods is shown in bold and other methods are marked with an asterisk if they are significantly worse than the best method according to a paired t-test with 5\% significance level. Lastly, the last column of the table, \emph{Selected-$\pi$}, indicates which method, among RNN-$\pi^{(1/2)}$, was selected as the best method on the validation set using MSE.

As one can note, except for \AEP\, where the baselines are better than RNN-based methods, for all other datasets, the best results are obtained with RNN-$\pi^{(1)}$ and RNN-$\pi^{(2)}$, these results being furthermore significantly better than the ones obtained with RNN-A and baseline methods, for both MSE and SMAPE. The MSE improvement varies from one dataset to another: between 8\% (\PW) and 26\% (\NAB) w.r.t RNN-A. Compared to ARIMA, one can achieve an improvement ranging from 18\% (15\%), in \OLD\,, to 94\% (75\%), in \PSE\,, w.r.t MSE (SMAPE). In addition, the selected RNN-$\pi$ method (column \emph{Selected-$\pi$}) is the best performing method on three out of six datasets (\AQ, \PW, \PSE) and the best performing RNN-$\pi$ method on \AEP. It is furthermore equivalent to the best performing method on \OLD, the only dataset on which the selection fails being \NAB\ (a failure means here that the selection does not select the best RNN-$\pi$ method). However, on this dataset, the selected method is still better than the original attention model and the baselines. Overall, these results show that RNN-$\pi^{(1/2)}$ significantly improves forecasting in the univariate time series we considered, and that one can automatically select the best RNN-$\pi$ method.

\begin{table}[t]
\centering
\caption{Overall results for univariate case with MSE (left value) and SMAPE (right value).}
\label{table:univar}
\begin{tabular}{|c|c|c|c|c|c||c|}
\hline
Dataset	&	RNN-A	&	RNN-$\pi^{(1)}$	&	RNN-$\pi^{(2)}$	&	ARIMA	&	RF	&	 \emph{Selected-$\pi$} \\ \hline 
\AQ	&	0.282$^*$/0.694$^*$	&	0.257/\textbf{0.661}	&	\textbf{0.25}/0.669	&	0.546$^*$/0.962$^*$	&	0.299$^*$/0.762$^*$	&	$\pi^{(2)}$ \\ \hline 
\OLD	&	0.319$^*$/0.595$^*$	&	\textbf{0.271}/\textbf{0.523}	&	0.275/0.586$^*$	&	0.331$^*$/0.619$^*$	&	0.305$^*$/0.606$^*$	&	$\pi^{(2)}$ \\ \hline
\AEP	&	0.025$^*$/0.085$^*$	&	0.029$^*$/0.101$^*$	&	0.027$^*$/0.095$^*$	&	\textbf{0.021}/\textbf{0.066}	&	0.021/0.085$^*$	&	$\pi^{(2)}$ \\ \hline 
\NAB	&	0.642$^*$/0.442$^*$	&	\textbf{0.475}/\textbf{0.323}	&	0.54$^*$/0.369$^*$	&	1.677$^*$/1.31$^*$	&	0.779$^*$/0.608$^*$	&	$\pi^{(2)}$ \\ \hline 
\PW	&	0.166$^*$/0.558	&	\textbf{0.152}/0.547	&	0.162$^*$/0.565$^*$	&	0.213$^*$/0.61$^*$	&	0.156/\textbf{0.544}	&	$\pi^{(1)}$ \\ \hline 
\PSE	&	0.034$^*$/0.282$^*$	&	\textbf{0.032}/0.264$^*$	&	0.033$^*$/\textbf{0.256}	&	0.623$^*$/1.006$^*$	&	0.053$^*$/0.318$^*$	&	$\pi^{(1)}$ \\ \hline 

\end{tabular}
\end{table}
 
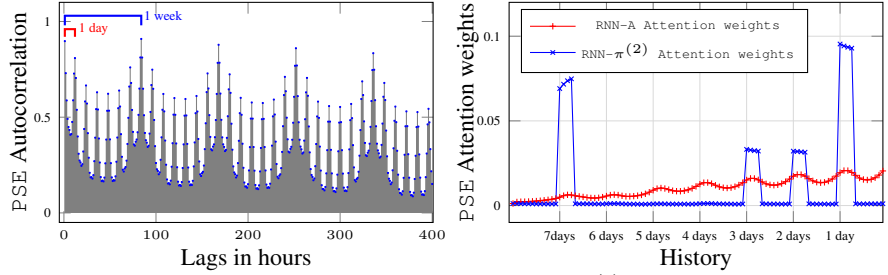
\begin{figure}[t]
	\centering     
	\pgfplotstableread{new_figs/pse_acf_data.txt} {\pseacf}
\pgfplotstableread{new_figs/attention_weights/attention_weights_pse_128_256_RNNA.txt} {\psevonetable}
\pgfplotstableread{new_figs/attention_weights/attention_weights_pse_128_256_RNNV2.txt} {\psevtwotable}
\begin{tikzpicture}
\begin{groupplot} [
    group style={group size= 2 by 1},
    height=4.5cm,width=6.5cm,
]
\nextgroupplot[
    ymin=-0.05, ymax=1.1,
    xlabel=Lags in hours,
    x label style={font=\small,at={(axis description cs:0.5,0.1)}},
    ylabel={\PSE } Autocorrelation,
    y label style={font=\small,at={(axis description cs:0.14,.5)}},
    xticklabel style = {font=\tiny},
    xmin=-5, xmax=401,
    ytick={0,0.5,1},
    yticklabels={0,0.5,1},
    yticklabel style = {font=\tiny},
]
\addplot [ycomb,color=gray] table [y={y}] {\pseacf};
\addplot [only marks,mark=*,mark options={scale=0.1},color=blue] table [y={y}] {\pseacf};
\draw[red,thick] (axis cs:1,0.92) -- (axis cs:1,0.96) -- node[anchor=west] {{\tiny 1 day}} (axis cs:12,0.96) -- (axis cs:12,0.92);
\draw[blue,thick] (axis cs:1,0.98) -- (axis cs:1,1.03) -- node[anchor=west,xshift=0.42cm] {{\tiny 1 week}} (axis cs:84,1.03) -- (axis cs:84,0.98);
\nextgroupplot[
    ymin=-0.01,ymax=0.12,
    grid=both,
    grid style={line width=.1pt, draw=gray!30},
    xlabel=History,
    x label style={font=\small,at={(axis description cs:0.5,0.1)}},
    ylabel={\PSE } Attention weights,
    y label style={font=\small,at={(axis description cs:0.14,.5)}},
    xticklabel style = {font=\tiny},
    xmin=0, xmax=96,
    xtick={13,25,37,49,61,73,85},
    xticklabels={7days, 6 days, 5 days, 4 days, 3 days, 2 days, 1 day},
    yticklabel style = {font=\tiny},
    ytick={0,0.05,0.1,0.15},
    yticklabels={0,0.05,0.1,0.15},
    legend style={at={(0.8,0.97)},font=\tiny,anchor=north east},
]
    \addplot [red,mark = +,mark options={scale=0.6}] table {\psevonetable};
    \addlegendentry{\tiny \texttt{RNN-A Attention weights}}
    \addplot [blue,mark = x, mark options={scale=0.6}] table {\psevtwotable};
    \addlegendentry{\tiny \texttt{RNN-$\pi^{(2)}$ Attention weights}}
\end{groupplot}
\end{tikzpicture} \vspace{-0.5cm}
        \caption{Autocorrelation of \PSE\, (left). RNN-A and RNN-$\pi^{(2)}$ attention weights (right).}
	\label{fig:attention_tau}
\end{figure}

Lastly, to illustrate the ability of RNN-$\pi$ to capture pseudo-periods, we display in Figure~\ref{fig:attention_tau} (left) the autocorrelation plot for \PSE, and in Figure~\ref{fig:attention_tau} (right) the average attention weights for the same time series obtained with RNN-A and RNN-$\pi^{(2)}$ (averaged over all test examples and forecast horizon points). As one can see from the autocorrelation plot, \PSE\ has two main weekly and daily pseudo-periods. This two pseudo-periods are clearly visible in the attention weights of RNN-$\pi^{(2)}$ that gives higher weights to the four points located at positions \textit{minus 7 days} and \textit{minus 1 day} (these four points correspond to the four points of the forecast horizon). The attention weights of RNN-$\pi^{(1)}$ (not shown here for space reasons) are very similar. In contrast, the attention weights of the original attention model (RNN-A) follow a general increasing behaviour with more weights on the more recent time stamps. This model thus misses the pseudo-periods.

\vspace{0.1cm}
\noindent \textbf{Results on multivariate time series}\label{subsec:multivar}
\vspace{0.1cm}

As mentioned in Table~\ref{tab:datasets}, we furthermore conducted multivariate experiments on \AQ, \AEP\ and \OLD\ using the multivariate extensions described in Section \ref{sec:model-univ}. For \AQ, we selected the four variables associated to real sensors, namely C6H6(GT), NO2(GT), CO(GT) and NOx(GT) and predicted the same one as the univariate case (C6H6(GT)). For \AEP, we selected two temperature time series, namely T1 and T6, and two humidity time series, RH6  and RH8, and we predict RH6 as in the univariate case. For \OLD, we trained the model using T0 to T3 and predicted T3, as we did on the univariate case. As RF outperformed ARIMA on five out of six univariate datasets and was equivalent on the sixth one, we retained only RF and RNN-A for comparison with RNN-$\pi^{(1/2/3)}$.

Table \ref{table:multivar} shows the results of our experiments on multivariate sets with MSE (SMAPE, not displayed here for readability reasons, has a similar behaviour). As before, for each time series, the best result is in bold and an asterisk indicates that the method is significantly worse than the best method (again according to a paired t-test with 5\% significance level). Similarly to the univariate case, the best results, that are always significantly better than the other results, are obtained with the RNN-$\pi$ methods: for \AQ\ and \OLD\, datasets, RNN-$\pi$ can respectively bring 24\% and 18\% of significant improvement over RNN-A. Similarly, for \AEP, the improvement is significant over RNN-A (17\% with RNN-$\pi^{(1)}$). Compared to RF, one can obtain between 11\% (\AEP) and 40\% (\AQ) of improvement.
As one can note, the selected method is always RNN-$\pi^{(3)}$. The selection is this time not as good as for the univariate case as the best method (sometimes significantly better than the one selected) is missed. That said, RNN-$\pi^{(3)}$ remains better than the state-of-the-art baselines retained, RNN-A and RF.

\begin{table}[t]
\centering
\caption{Overall results for multivariate case with MSE.}
\label{table:multivar}
\begin{tabular}{|c|c|c|c|c|c||c|}
\hline
Dataset	&	RNN-A	&	RNN-$\pi^{(1)}$	&	RNN-$\pi^{(2)}$	&	RNN-$\pi^{(3)}$	&	RF	&	 \emph{Selected-$\pi$} \\ \hline 
\AQ	&	0.352$^*$	&	0.276$^*$	&	\textbf{0.268}	&	0.3$^*$	&	0.45$^*$	&	$\pi^{(3)}$ \\ \hline 
\OLD	&	0.336$^*$	&	0.328$^*$	&	0.327$^*$	&	\textbf{0.274}	&	0.315$^*$	&	$\pi^{(3)}$ \\ \hline 
\AEP	&	0.029$^*$	&	\textbf{0.024}	&	0.036$^*$	&	0.026$^*$	&	0.027$^*$	&	$\pi^{(3)}$ \\ \hline 

\end{tabular}
\end{table}

\vspace{-0.4cm}
\vspace{-0.1cm}
\section{Conclusion}
\label{sec:conclusion}
\vspace{-0.2cm}

We studied in this paper the use of Seq-RNNs, in particular the state-of-the-art bidirectional LSTMs encoder-decoder with a content attention model, for modelling and forecasting time series. If content attention models are crucial for this task, they were not designed for time series and currently are deficient as they do not capture pseudo-periods. We thus proposed three extensions of the content attention model making use of the (relative) positions in the input and output sequences (hence the term \textit{position-based content attention}). The experiments we conducted over several univariate and multivariate time series demonstrate the effectiveness of these extensions, on time series with either clear pseudo-periods, as \PSE, or less clear ones, as \AEP. Indeed, these extensions perform significantly better than the original attention model as well as state-of-the-art baseline methods based on ARIMA and random forests.

In the future, we plan on studying formal criteria to select the best extension for both univariate and multivariate time series. This would allow one to avoid using a validation set that may be not large enough to properly select the best method. We conjecture that this is what happening on the multivariate time series we have retained.

\vspace{-0.3cm}

\bibliographystyle{splncs04}
\bibliography{main}

\end{document}